\documentclass[10pt,twocolumn,letterpaper]{article}

\usepackage{cvpr}              % To produce the CAMERA-READY version
\usepackage{amsmath}
\usepackage{graphicx}
\usepackage{multirow}
\usepackage{amssymb}
\usepackage[pagebackref,breaklinks,colorlinks]{hyperref}

\usepackage[capitalize]{cleveref}
\crefname{section}{Sec.}{Secs.}
\Crefname{section}{Section}{Sections}
\Crefname{table}{Table}{Tables}
\crefname{table}{Tab.}{Tabs.}

\def\cvprPaperID{11338} % *** Enter the CVPR Paper ID here
\def\confName{CVPR}
\def\confYear{2023}
\begin{document}

\title{Deformable Kernel Expansion Model for Efficient Arbitrary-shaped Scene Text Detection}

\author{Tao He\\
% Chongqing University \\
{\tt\small taohe@cqu.edu.cn}
\and
Sheng Huang\\
{\tt\small huangsheng@cqu.edu.cn}
\and
Wenhao Tang\\
{\tt\small whtang@cqu.edu.cn}
\and    
Bo Liu\\
{\tt\small kfliubo@gmail.com}
}

\maketitle

%%%%%%%%% ABSTRACT
\begin{abstract}
Scene text detection is a challenging computer vision task due to the high variation in text shapes and ratios. In this work, we propose a scene text detector named Deformable Kernel Expansion (DKE), which incorporates the merits of both segmentation and contour-based detectors. DKE employs a segmentation module to segment the shrunken text region as the text kernel, then expands the text kernel contour to obtain text boundary by regressing the vertex-wise offsets. Generating the text kernel by segmentation enables DKE to inherit the arbitrary-shaped text region modeling capability of segmentation-based detectors. Regressing the kernel contour with some sampled vertices enables DKE to avoid the complicated pixel-level post-processing and better learn contour deformation as the contour-based detectors. Moreover, we propose an Optimal Bipartite Graph Matching Loss (OBGML) that measures the matching error between the predicted contour and the ground truth, which efficiently minimizes the global contour matching distance. Extensive experiments on CTW1500, Total-Text, MSRA-TD500, and ICDAR2015 demonstrate that DKE achieves a good tradeoff between accuracy and efficiency in scene text detection.
    %Detecting text instances from real-world scenes remains a grand challenge due to the variation in shapes and ratios. In this work, we propose a novel and efficient text detector named Deformable Kernel Expansion (DKE) for detecting arbitrary-shaped text, which directly regresses accurate text boundaries from kernels produced by segmentation results. Our method consists of Text Kernel Generation (TKG) module and Deformable Contour Expansion
 %(DCE) module. A lightweight segmentation network constructs the TKG to get pixel-level segmentation results of shrunk text areas. 
 %To avoid the complicated pixel-level post-processing, we treat those process as a contour deformaion task and deploy contour deformation module DKE to expand the shrunk text areas to text boundaries in contour forms. Moreover, to facilitate the kernel expansion procedure, we match the predicted and ground truth vertex pairs by minimizing the total distance of deforming predicted contours to annotated boundaries and propose the corresponding loss function named optimal bipartite graph matching loss.
 %Experiments conducted on CTW1500, Total-Text, MSRA-TD500, and ICDAR2015 have demonstrated the proposed method excels in detecting arbitrary-shape texts while maintaining real-time inference speed.
 \end{abstract}
 
 \begin{figure}[t] %插入图片
    \centering %图片居中
    \includegraphics[width=7.5cm]{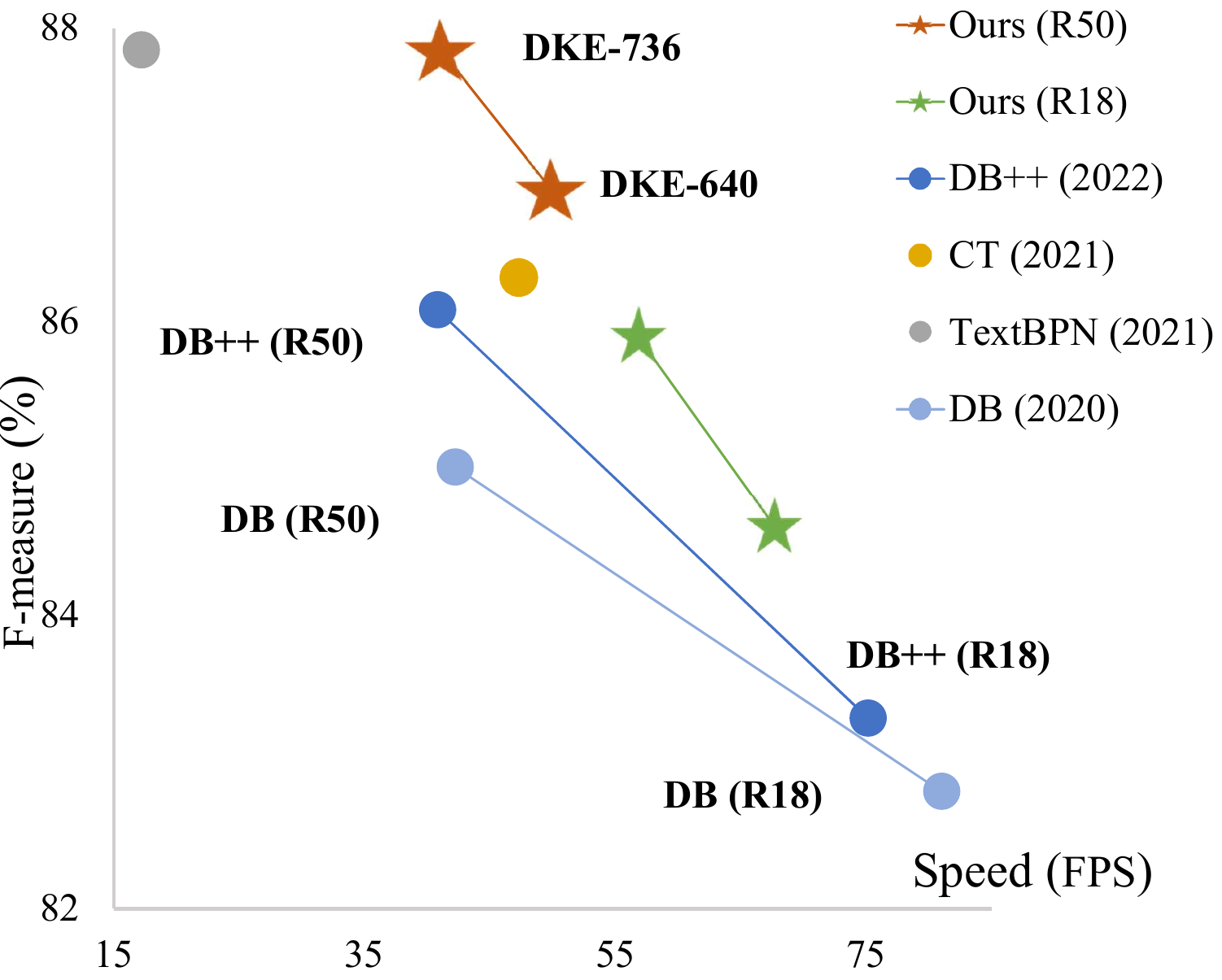} %[图片大小]{图片路径}
    \caption{The accuracy and inference speeds of several top-performance scene text detectors on the Total-Text dataset, where DB, R50 and R18 are DBNet, ResNet50 and ResNet18 respectively. 
    %In testing our method, the short side of the test image is resized to different scales (\eg, 736 and 640) while keeping the original aspect ratio.
     %DB = DBNet, R50, R18 denote ResNet50 and ResNet18, respectively.
     All models are tested as a Two-Step text detector, which outputs polygons as their final results.
    For more details, please see Sec.\ref{details}.}
    \label{Figure:speed_table}
    \vspace{-0.5cm}
 \end{figure}
 %%%%%%%%% BODY TEXT
 \section{Introduction}
 
 \label{sec:intro}
 Scene text detection (STD)~\cite{Scenetextdetection} has received increasing attention from academia and industry for wide applications such as intelligent transportation, cross-language information retrieval, visual search, and blind auxiliary. Precisely localizing the region of text instances in natural images is essential for improving or achieving such applications. With the development of text detection methods, how to detect text instances in scene images accurately and efficiently remains a significant challenge for researchers.
 
 The significant challenges in STD are handling the variety of text shapes, scales, and extreme aspect ratios of text regions. Segmentation-based detectors~\cite{PSE,SAE,PAN,wang2021pan++,db,db++}) 
 output pixel-wise predictions based on local texture, which is proven effective to the above challenges. Compared with other approaches (\eg, regression-based and contour-based methods), they possess great advantages in speed while maintaining similar accuracy. 
 However, the local nature 
 makes segmentation-based method reply on complicated post-processing to obtain the text boundary. 
 For example, TextSnake~\cite{textsnake} proposes a striding algorithm which does vertex-wise boundary detection along the central line of the response map. PSENet~\cite{PSE}
 generates multi-scale kernel representations of text regions and adopts a progressive BFS-based scale expansion algorithm to merge regions at the pixel level. Several recent segmentation-based STD models improve the inference efficiency by simplifying the post-processing, such as the differentiable binarization model~\cite{db++}. However, their accuracy heavily relies on the segmentation results. The lack of contour adjustment mechanisms leads to ineffective detection if the segmentation has a large deviation.
 
  Motivated by contour-based methods~\cite{CurveGCN,DeepSnake,Dance,TaoZhang2022E2ECAE} in instance segmentation, another branch of STD models discrete the text contour as a series of discrete vertices and directly predicts the coordinates of those vertices. The adaptive text region representation model~\cite{ATTR} uses a text region proposal network to obtain text region proposals. Then RNN model is learned to predict the coordinates of boundary vertices as a sequence. The TextBPN model~\cite{TextBPN} learns an adaptive deformation model for text boundary vertices adjusting. Although achieving impressive boundary accuracy, this type of model usually needs to do vertex-wise adjusting iteratively or sequentially, leading to an inefficient inference process.

 In this work, we propose a novel STD model named Deformable Kernel Expansion (DKE) which leverages the merits of both segmentation and contour-based STD methods.
 Specifically, we extract dense contour vertices from the shrunken text kernel and the annotated text boundary for Deformable Contour Expansion (DCE) module learning. This methodology is different from most contour-based STD models that adjust coarse text boundaries to get more accurate text boundaries, which is more efficient. Because the text kernel, as a more centralized text region, has less noise than text boundaries. 
 One technical issue of contour expansion is to find the vertex pairing relations between predicted vertices and annotated boundaries for a contour deformation loss calculation. Existing contour-based models usually pair them in a fixed manner, regardless of the continuous position adjustment of the predicted vertex. We propose a vertex pairing strategy based on minimizing the deformation cost from a global perspective and present a novel contour deformation loss named the Optimal Bipartite Graph Matching Loss (OBGML).  
 Benefiting from DKE and OBGML, our detector achieves superior or competitive results compared to others on the curved and oriented text benchmarks. As shown in Fig.\ref{Figure:speed_table}, we achieve the best tradeoff between accuracy and efficiency compared with recent top-performance scene text detectors on detecting arbitrary shape texts.
 The technical contribution of the proposed model can be summarized as follows: 
 \begin{itemize}
 % namely Text Kernel Generation (TKG) and Deformable Contour Expansion (DCE).  
    \item[$\bullet$] We propose an efficient scene text detector named DKE. After the initial text segmentation and text kernel extraction, post-processing is performed by a learnable regression network. The regression network deforms the text kernel via contour deformation in one step, making the generation of text boundary fast and accurate.
 
    \item[$\bullet$]  We propose to use the contour of the text kernel instead of the text region to initialize the scene text contour deformation. Such initialization is less affected by the noise while enabling the deformation to be accomplished in a single iteration.
    % Through experiments, we have proved the feasibility and effectiveness of using \emph{contour deformaion} to expand text kernels to text boundaries in forms of vertexes.
    % in relative long distance. 
    % We proposed an efficient text detector named STD for detecting arbitrary-shaped texts which adopted circular convolution to regress text boundaries from its kernels. No extra post-processing is required for our methods except for fitting polygons with fewer points.
    % in order to balance between accuracy and inference speed. 
    % \item[$\bullet$]We design a loss function $L_{otc}$ based on minimizing the total transportation cost to encourage the model to find the ideal deformation path for transforming kernals to boundaries. 
    \item[$\bullet$] 
    % We propose a optimal transport based contour point matching strategy, which is more effect to find the  
    % The existing deformation strategies for \emph{contour deformation} are not optimal in solving expansion problem. 
    We formulate the contour vertices pairing between prediction and the ground truth as a bipartite graph matching problem for solution and define a novel contour deformation loss named OBGML. It enables minimizing the total distance of deforming predicted contours to annotated boundaries globally and further improving the performances of DKE. 
    % It enables to globally minimize the total distance of deforming predicted contours to annotated boundaries and further improve the performances of DKE.
 \end{itemize}
 
 \section{Related Work}
 Scene text detection methods have developed rapidly
 in recent years. In the era of deep learning, there exist
 two popular branches: segmentation-based methods and
 contour-based methods. They are widely studied for their
 ability to represent text instances with arbitrary shapes.
 
 \textbf{Segmentation-based Detector: }
 Segmentation-based methods\cite{textsnake,Masktextspotter,MaskTextSpotterv3,PSE,SAE,PAN,db,db++} often predict text regions, or basic components of scene texts, such as text kernels~\cite{PSE,SAE,PAN,db}, text center region~\cite{textsnake}, to group pixels into different text instances. 
 The most significant advantage of segmentation-based methods is the flexible representation of text instances 
 with arbitrary shapes. The pipeline of those methods originated from instance segmentation methods. Inherited from the instance segmentation method Mask R-CNN~\cite{maskrcnn}, Mask TextSpotter~\cite{Masktextspotter} detected scene texts by instance segmentation which enables it to detect arbitrary-shaped text. To effectively separate adjacent texts, PSENet~\cite{PSE} adopted a progressive scale algorithm to gradually expand the pre-defined text kernels.
 PAN~\cite{PAN} replaced the post-processing of PSENet with a learnable module to predict similarity vectors of pixels to achieve better performance. DB~\cite{db} simplified the post-processing by directly expanding the kernel contours to speed up the detection. However, the lack of contour adjustment mechanisms makes DB\cite{db} vulnerable to the deviation of segmentation results.
 
 \begin{figure*}[t]
 \begin{center}
 \includegraphics[width=15cm]{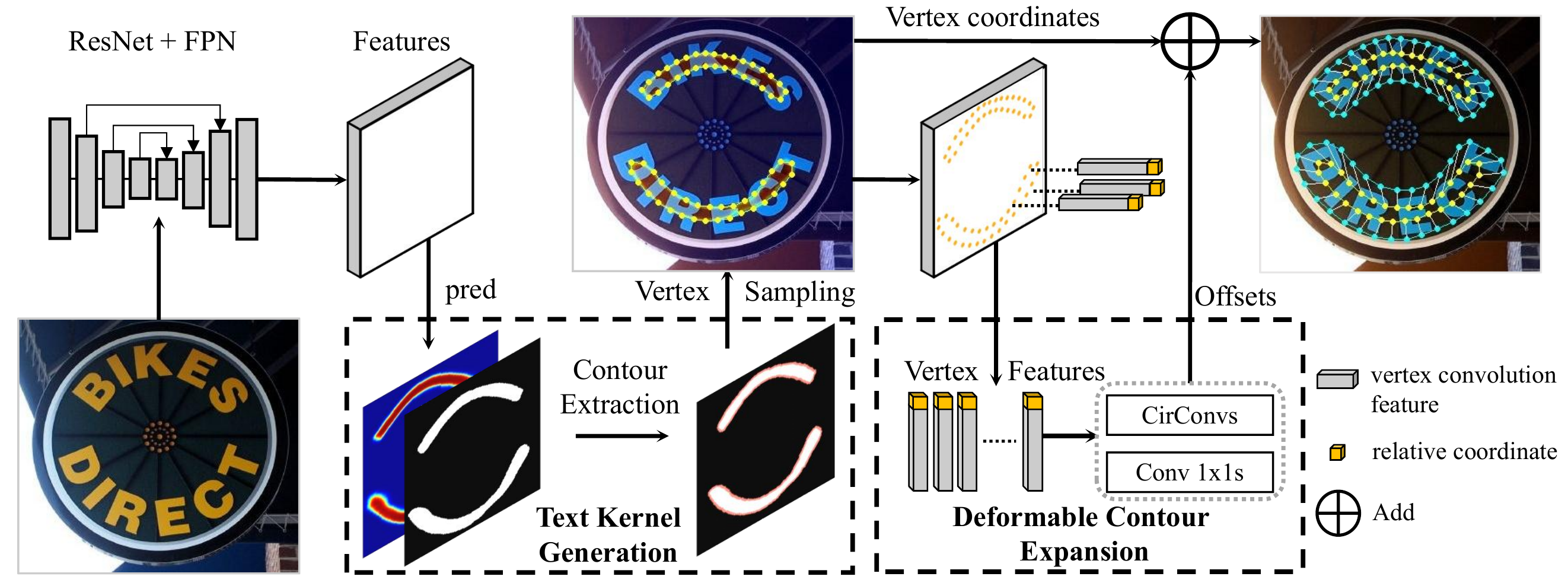} %[图片大小]{图片路径}
 \vspace{-0.3cm}
 \end{center}
    \caption{The overall architecture of our network. Our network consists of a backbone network to extract image features, an FPN to fuse the multi-scale features, a \textbf{Text Kernel Generation} module to detect text kernels at the pixel level, and a \textbf{Deformable Contour Expansion} module to deform kernels to text boundaries. Specifically, multi-scale feature maps are first produced from the backbone network. Next, multi-scale feature maps are fused by an FPN network to get the contextual feature. Then, a segmentation-based detector is used to predict the probability map of the text kernel. Then we uniformly sample $N$ vertices on the kernel contour. After that, a feature sampling procedure is conducted to yield the vertex features by concatenating the local CNN features and relative coordinates of each vertex. Those features are employed to predicted vertex-wise offsets. Finally, the text boundary can be calculated by adding the sample $N$ vertex coordinates with corresponding offsets.
  }
 \label{Architecture}
  \vspace{-0.4cm}
 \end{figure*}   
 
 \textbf{Contour-based Detector: }
 Contour-based methods~\cite{ATTR,weaklyarbitrary,contournet,dai2021progressive,TextBPN} deem scene text detection as a regression task. Unlike many regression-based methods~\cite{East,Textboxes,deepseg,liu2020abcnet}, those methods usually directly predict vertices on text boundaries.
 ContourNet~\cite{contournet} used a series of discrete vertices to represent text boundaries. It adopted a local orthogonal texture-aware module to model the local texture information of text proposals and then predicted vertices on text region boundaries. 
 TextBPN~\cite{TextBPN} generated boundary proposals for text regions by a boundary proposal model and 
 then adopted an adaptive boundary deformation model to refine the coarse boundaries iteratively.
 Although each iteration of contour refining is quite efficient, the contour-based methods often need to refine the coarse contour several times, which slows down the overall inference speed. Generally, the contour-based methods have a gap with the segmentation-based methods in inference speed.
 
 In real-world applications, a good scene text detector should be not only accurate but efficiency. Some works consider both of these two aspects to elaborate scene text detectors.
 PAN~\cite{PAN} is proposed to adopt a low computational-cost segmentation head and learnable post-processing for scene text detection, which achieves a good balance between accuracy and efficient. 
 CT~\cite{CentripetalText} reconstructs text boundaries using heuristics based on text kernels and centripetal shifts. This process can be calculated in parallel by implementing one matrix operation, guaranteeing good efficiency without losing accuracy. Even though they are considerably fast in clustering text instances at the pixel level,  the contour extraction procedure takes more than half of their inference time. This makes them not suitable for applications that require text boundaries as outputs. 
 
 Our method incorporates both segmentation-based and contour-based methods. It employs the segmentation-based methodology for generating text kernels as high-quality initial detection boundaries while regressing the final detection contour via a contour deformation module. Therefore, our method inherits the good text representation ability of the segmentation-based methodology and the contour deformation efficiency of contour-based methodology in each iteration. In other words, our method can introduce a good tradeoff between accuracy and efficiency.

 \section{Methodology}
 The Deformable Kernel Expansion (DKE) model consists of two core modules, namely Text Kernel Generation (TKG) and Deformable Contour Expansion (DCE).
 The TKG step aims at generating a high-quality text kernel to provide a good initial boundary,
 while the DCE step aims at learning to expand the kernel contours for obtaining the final detection.
 \subsection{Network Architecture}
 The entire architecture of our model is summarized as Fig.\ref{Architecture}.
 Following the conventions~\cite{PSE,SAE,PAN,db}, the ResNet~\cite{resnet} is employed to extract multi-scale features,
 and then these features are fused as the contextual feature $f$ by a feature-pyramid network~\cite{FPN}.
 This step can be mathematically denoted as $f=F(I)$, where $F(\cdot)$ is the hybrid mapping of the
 ResNet and the feature-pyramid network, and $I$ is an input image. 
 
 In the TKG step,
 we follow the pipeline of segmentation-based scene text detection methods, and leverage a lightweight
 segmentation network $S(\cdot)$ as an initial detection head to predict the probability map $\hat{Y}=S(f)$,
 which encodes text kernel $\mathcal{K}$ at the pixel level.
 We uniformly sample $N$ vertices on the kernel boundary as the contour representation, which is denoted as $P^{\mathcal{K}}=\{p_i:=(x_i,y_j)\}^{N}_{i=1}$, and used as the initial detection boundary.
 A contour composed of $N$ (\eg, $N = 128$) vertices is sufficient to describe most of the instances well~\cite{DeepSnake}.
 % $N$ is empirically set to 128.
 Moreover,
 these vertices are ordered in certain rules, which will be introduced in Sec.\ref{Label_generation}.
 By default, $P^{\mathcal{K}}$ mentioned in our literature is the re-ranked vertex collection.
 
 In the DCE step, we obtain the convolution features for each vertex in $P^{\mathcal{K}}$ by retrieving them in the feature $f$ based on its coordinates.
 The previous method~\cite{failureCOORDCONV} indicates that learning the coordinate mapping between Cartesian space and the pixel space is challenging. Therefore, we follow some recent works~\cite{XinlongWang2020SOLOv2DA,liu2021abcnetv2},
 which append the relative coordinates of vertices at the end of convolution features as extra information cues, to alleviate this issue. In our method, the upper left corner of the minimum bounding box is considered the origin of the coordinate system for calculating the relative coordinates. The aforementioned hybrid feature processing is formulated as follows,
 \begin{equation}
 U=\omega(P^{\mathcal{K}},f) =\omega(\{p_i\}^{N}_{i=1},f),
 \end{equation}
 where $U:=\{u_i\}_{i=1}^N$ are the obtained vertex features and $\omega(\cdot,\cdot)$ is the mapping function of the aforementioned process.
 
 The goal of the DCE step is to learn the mapping between these vertex representations
 and the offsets of vertices on the kernel contour to the ones on the boundary of the final detection box with a contour deformation network\cite{DeepSnake} $D(\cdot)$,
 \begin{equation}\label{offset}
    \triangle G^{\mathcal{K}}:= \{(\triangle x_i, \triangle y_i)\}_{i=1}^N = D(U)=D(\{u_i\}_{i=1}^N),
 \end{equation}
 where $\triangle G^{\mathcal{K}}$ are the learned coordinate offsets of $P^{\mathcal{K}}$ to the ground truth $G^\mathcal{K}$. Then the predicted coordinates of the corresponding vertices on the final bounding box can be obtained as follows,
 \begin{equation}
 \hat{G}^{\mathcal{K}}=P^{\mathcal{K}}+\triangle G^{\mathcal{K}}.
 \end{equation}
 We adopt the small text kernel, because it is easier to separate two adjacent text regions and less likely disturbed by the background. Therefore, the contour deformation process is actually a kernel expansion process.
 
 \subsection{Kernel Annotation and Vertex Sorting}
 \label{Label_generation}
 In the training phase, we need to generate the text kernel and the contour ground truth for each training example based on its annotated text boundary. We follow the previous works~\cite{PSE,db}, and shrink the annotated text boundary into the text kernel via the Vatti clipping algorithm~\cite{BalaRVatti1992AGS}. This algorithm is able to calculate the shrinking margin $m$ between the shrunken kernel contour and the original text boundary based on a given shrink ratio. In our experiments, we empirically set the shrink ratio to 0.4 for generating the small text kernel.
  \begin{figure}[b] %插入图片
    \centering %图片居中
    \vspace{-0.4cm}
    \includegraphics[width=8cm]{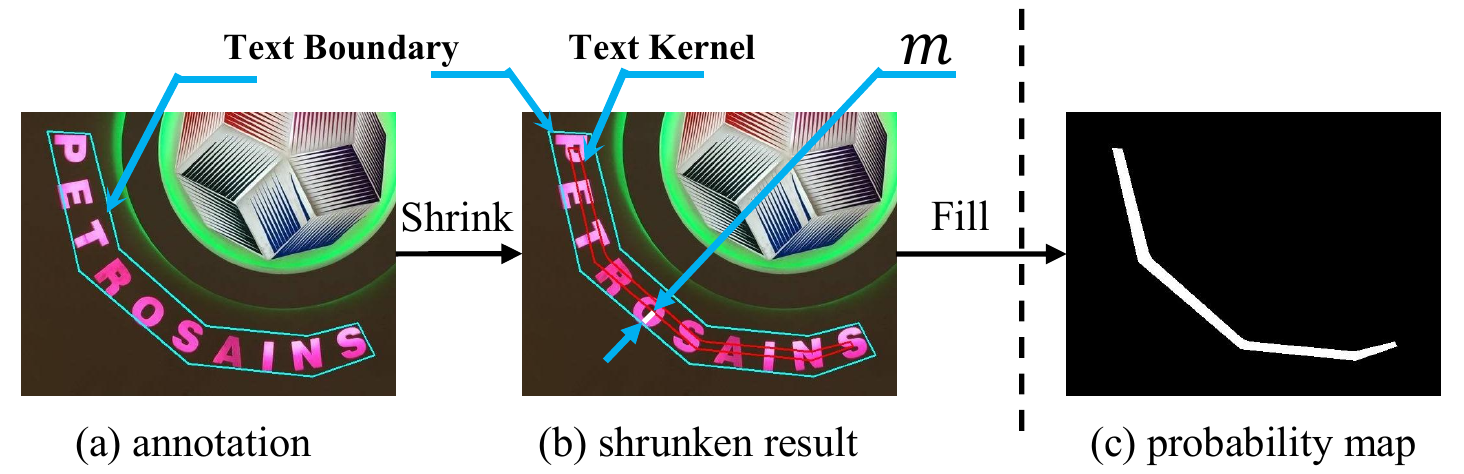} %[图片大小]{图片路径}
    \vspace{-0.1cm}
    \caption{Illustration of the text kernel annotation. (a) The annotated text boundary. (b) The contour of the text kernel and its shrinking margin $m$ calculated by the shrinking algorithm. (c) The generated probability map for training the DCE module. }
    \label{Figure:label_generation}
    \vspace{-0.2cm}
 \end{figure}

 After that, we uniformly sample $N$ vertices on the generated text kernel boundary for depicting the kernel contour. These sampled vertices are sorted in a clockwise manner, and the vertex that is the closest point to the upper left corner of the minimum surrounding rectangle of the text kernel is deemed as the first vertex.
 
 Let $\Phi(\cdot)$ be such a vertex sample and sorting operation. The kernel boundary can be denoted as a sorted vertex collection as mentioned, $P^{\mathcal{K}}=\{p_i:=(x_i,y_j)\}^{N}_{i=1}=\Phi(\tilde{Y})$, where $\tilde{Y}$ can be the text kernel of a testing example generated by the segmentation network $\hat{Y}$ or the pre-processed text kernel of a training example (the ground truth) $Y$.

 %In the training phase, we need to generate the text kernel label and the contour representation for each training example based on its annotated text boundary. As shown in Fig.\ref{Figure:label_generation}, we follow the previous works~\cite{PSE,db}, and shrink the annotated text boundary into the text kernel via Vatti clipping algorithm~\cite{BalaRVatti1992AGS}. This algorithm is able to calculate the shrinking margin $m$ between the shrunk kernel contour and the annotated text boundary based on a given shrinking ratio. In our experiments, we empirically set the shrinking ratio to 0.4 for generating the small text kernel.
 
 %Then, we uniformly sample $N$ vertices on the generated text kernel contour for depicting the contour representation. These sampled vertices are sorted in clockwise manner, and the vertex, which is the closest point to the upper left corner of the minimum surrounding rectangle of the text kernel, is deemed as the first vertex.
 
 %Let $\Phi(\cdot)$ be such a vertex sample and sorting operation. The kernel contour can be denoted as a sorted vertex collection as mentioned, $P^{\mathcal{K}}=\{p_i:=(x_i,y_j)\}^{N}_{i=1}=\Phi(\tilde{Y})$, where $\tilde{Y}$ can be the text kernel of a testing example generated by the segmentation network $\hat{Y}$ or the pre-processed text kernel of a training example (the ground truth) $Y$.

 \subsection{Model Optimization}
 
 The Deformable Kernel Expansion (DKE) model accomplishes accurate text detection from two aspects. On the one hand, DKE generates text kernels for an input image via a segmentation network. The kernels are used to capture the shape characteristics of text regions. On the other hand, a regression network is applied to learn the contour deformation of text kernels to the final detection boundaries only with several sampled vertices. In such a manner, the DKE model should simultaneously minimize the text kernel generation loss and the deformable contour expansion loss to achieve the above goals.
 
 \subsubsection{Text Kernel Generation} 
 %The text kernel generation is accomplished as a segmentation task, which is essentially a pixel binary classification problem. 
 In this step, the feature of a scene text image is fed into a segmentation-based detection head to obtain the pixel classification probability map $\hat{Y}=S(f)$, and then we use the binary cross-entropy for measuring the discrepancy between this generated kernel and its ground truth. The loss of this process can be mathematically denoted as follows,
 \begin{equation}\label{ce}
     \mathcal{L}_{s} = -\sum_{i\in \mathcal{A}} Y_i \log \hat{Y}_i+ (1-Y_i) \log (1-\hat{Y}_i),
 \end{equation}
 where $\hat{Y}_i$ is the predicted probability corresponding to the $i$-the pixel and $Y_i$ is its ground truth. Note, not all pixels will take part in the loss calculation. We only measure the prediction discrepancies of pixels in a pixel sub-set $\mathcal{A}$. We follow the previous works~\cite{db,CentripetalText} for applying this trick to alleviate the extreme imbalance between the positive and negative pixels where the pixels in the text region are deemed as positives while the reminder is regarded as negatives. In $\mathcal{A}$, all the positives will be preserved, while only some of negatives will be collected via a hard negative mining strategy, which will keep the ratio of positive to negative being 1:3.
 
 \begin{figure*}[t] %插入图片
    \centering %图片居中
    \includegraphics[width=13cm]{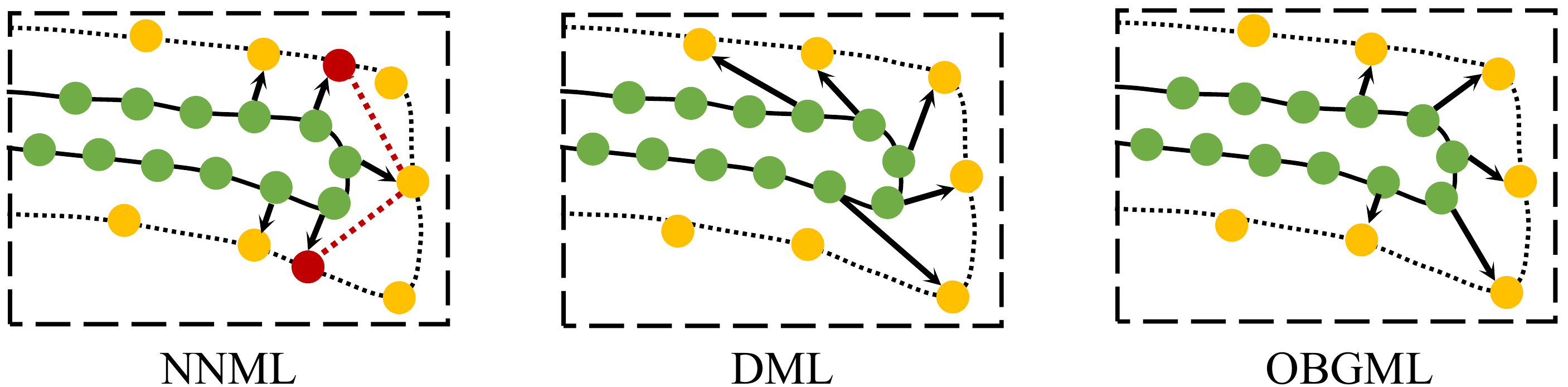} 
    \vspace{-0.1cm}
    \caption{Illustrations of the contour deformation paths of different losses. The \textcolor[RGB]{112,173,71}{green points} are the predicted contour vertices, the \textcolor[RGB]{255,192,0}{yellow points} are the labeled vertices, the \textcolor{red}{red points} are the newly sampled points on the annotated boundary, and the arrows represent the deformation path (the relation of vertex pairing). \textbf{In the best case, the vertices should be adjusted to the target contour with the least cost while maintaining complete structure as labeled vertices}. }
    \label{loss_show}
    \vspace{-0.4cm}
 \end{figure*}

 \subsubsection{Deformable Contour Expansion}
 The Deformable Contour Expansion is able to be achieved by predicting the offsets of vertices on the text kernel contour to the corresponding ones on the text boundary with a regression network~\cite{DeepSnake} as depicted in Equation~\ref{offset}. Note, in the training phase, the generated text kernel from annotations is leveraged for learning the contour deformation.
 
 Most contour-based methods conduct the vertex pairing in two strategies: direct matching~\cite{weaklyarbitrary,DeepSnake,TextBPN} and nearest neighbor matching~\cite{TaoZhang2022E2ECAE}. The first strategy does not conduct any vertex alignment and directly applies smooth $L_1$ loss to measure the pre-fixed vertex sequence discrepancy. The previous work~\cite{TaoZhang2022E2ECAE} indicates that the absence of vertex alignment will bring problems like slow convergence and even wrong prediction. They suggest aligning the predicted vertex sequence with the ground truth through a nearest-matching principle. To distinguish the losses of these two strategies, we name the contour deformation losses of these two approaches Direct Matching Loss (DML) and Nearest Neighbor Matching Loss (NNML), respectively.
  
 NNML essentially employs the greedy rule to find the correspondence of the predicted vertex in the ground truth, which cannot guarantee the optimal vertices pairing for kernel expansion process. Moreover, it cannot make sure all the vertices in the ground truth can be used in the optimization for accomplishing the expansion task, as shown in Figure~\ref{loss_show}.
 To address the above issues, we deem the vertex pairing problem as a bipartite graph matching problem for solution, and propose the \textbf{Optimal Bipartite Graph Matching Loss (OBGML)} for optimizing the deformable contour expansion. 
 
 The goal of the vertex pairing is to find a unique vertex in the ground truth as the correspondence for each predicted vertex that the total squared Euclidean distances between the predicted vertices and their correspondences are minimized.
 % We adopt squared Euclidean distance as cost to better suppress pairings with long distance.
 From the perspective of graph theory, the predicted vertices and their ground truths are two disjoint vertex sets, which can be depicted as a bipartite graph. Therefore, the aforementioned vertex pairing issue is actually a bipartite graph matching problem. This problem can be solved by the Hungary algorithm~\cite{kuhn1955hungarian}, and the global optimal solution enables to be obtained. A $N\times N$-dimensional vertex-vertex incidence matrix $M$ is constructed for encoding such a bipartite graph. The $ij$-th element of this matrix is the squared Euclidean distance between the $i$-th predicted vertices and the $j$-th real vertex,
 \begin{equation}
     M_{ij} = ||\hat{G}_i^{\mathcal{K}}-{G}_j^{\mathcal{K}}||_2^2.
     \label{cost}
 \end{equation}
 Then, the Hungary algorithm\cite{kuhn1955hungarian} is employed to find the best bipartite graph matching based on this incidence matrix,
 \begin{equation}
 % x_{i}^{*} = Hungarian(p_{i}^{kernal},p_{i}^{gt})
 \mathcal{H}:=\{(i,j^*)\}_{i=1}^N\in \mathcal{R}^{N\times2} = {\rm Hungarian}(M),
 \label{Hungarian}
 \end{equation}
 where $\mathcal{H}$ is the obtained index set of predicted vertices and their correspondences. 
 %Note, the Hungary algorithm enables to obtain the global optimal solution, and thereby guarantees the optimal vertex pairing. 
 Finally, the OBGML is formulated as follows,
 \begin{equation}
     \mathcal{L}_{r}=\frac{1}{N}\sum_{i=1}^{N} \rm smooth_{L_1}(\hat{G}_i^{\mathcal{K}}- G_{j^*}^{\mathcal{K}}).
     \label{oml}
 \end{equation}
 
 In such a manner, the DKE model can be obtained via minimizing the aforementioned two losses,
 \begin{equation}
   \{\hat{F},\hat{S},\hat{D}\}  \leftarrow\arg\underset{F,S,D}\min~\mathcal{L}:=\mathcal{L}_s+\lambda \mathcal{L}_{r},
 \end{equation}
 where $\lambda$ is a tunable hyper-parameter for reconciling two losses. According to the numeric values of the losses, $\lambda$ is set to 0.25 in all of our experiments.

 \subsection{Inference}
 In the inference phase, we can use the obtained networks $\hat{F}$,$\hat{S}$,and $\hat{D}$ for extracting the features, generating the text kernels and predicting the contour deformation offsets for an image $I$ as follows,
 \begin{equation}
  \triangle G^{\mathcal{K}} = \hat{D}(\omega(\phi(\hat{S}(\hat{F}(I))),\hat{F}(I))).
 \end{equation}
 Then, the final detection boundary depicted by $N$ vertices is produced as follows,
 \begin{equation}
 \hat{G}^{\mathcal{K}}:=\{(\hat{x}_i,\hat{y}_i)\}_{i=1}^N=\phi(\hat{S}(\hat{F}(I)))+\triangle G^{\mathcal{K}}.
 \end{equation}
 
 \section{Experiments}
    %  In this section, we first introduce the used datasets and implementation details in all our experiments.
    %  Then, we present ablation studies in several settings.
    %  Finally, we compare our proposed method with recent text detection methods in terms of accuracy and time-consuming
    %  during inference.
 \subsection{Datasets}
 % \subsubsection{Quadrangular Text}
 \textbf{SynthText}~\cite{AnkushGupta2016SyntheticDF} is a synthetic dataset consisting of more than 800$k$ synthetic images. It is only used to pre-train our model.
 
 \textbf{MLT-2017}\cite{MLT2017} is a multi-lingual dataset that contains 7200 training images, 1800 validation images, and 9000 test images. Those images are annotated with quadrilateral boxes at the word level. The training and validation images are used to pre-train our model.
 
 \textbf{ICDAR2015}~\cite{DimosthenisKaratzas2015ICDAR2C} dataset consists of 1000 training images and 500 testing images, which are captured by Google Glass. Most of the including text instances are severely distorted or blurred. The text instances are labeled at the word level with quadrilateral boxes.
 
 \textbf{MSRA-TD500}~\cite{msra-td500} dataset is a multi-language dataset that includes English and Chinese text instances. There are 300 training images and 200 testing images. The text instances are all labeled at the text-line level.
 
 % \subsubsection{Curved Text}
 \textbf{Total-Text}~\cite{CheeKhengChng2017TotalTextAC} is a curved text dataset including 1255 training and 300 testing images. It contains horizontal, multi-oriented, and curve text instances labeled at the word level.
 
 \textbf{CTW1500}~\cite{CTW1500} is another curved text dataset, including 1000 training images and 500 testing images. It contains both English and Chinese texts annotated at the text-line level with polygons.
 
 \begin{figure*}[t]
   \begin{center}
   % \fbox{\rule{0pt}{2in}
   % \rule{.9\linewidth}{0pt}}
   \includegraphics[width=14cm]{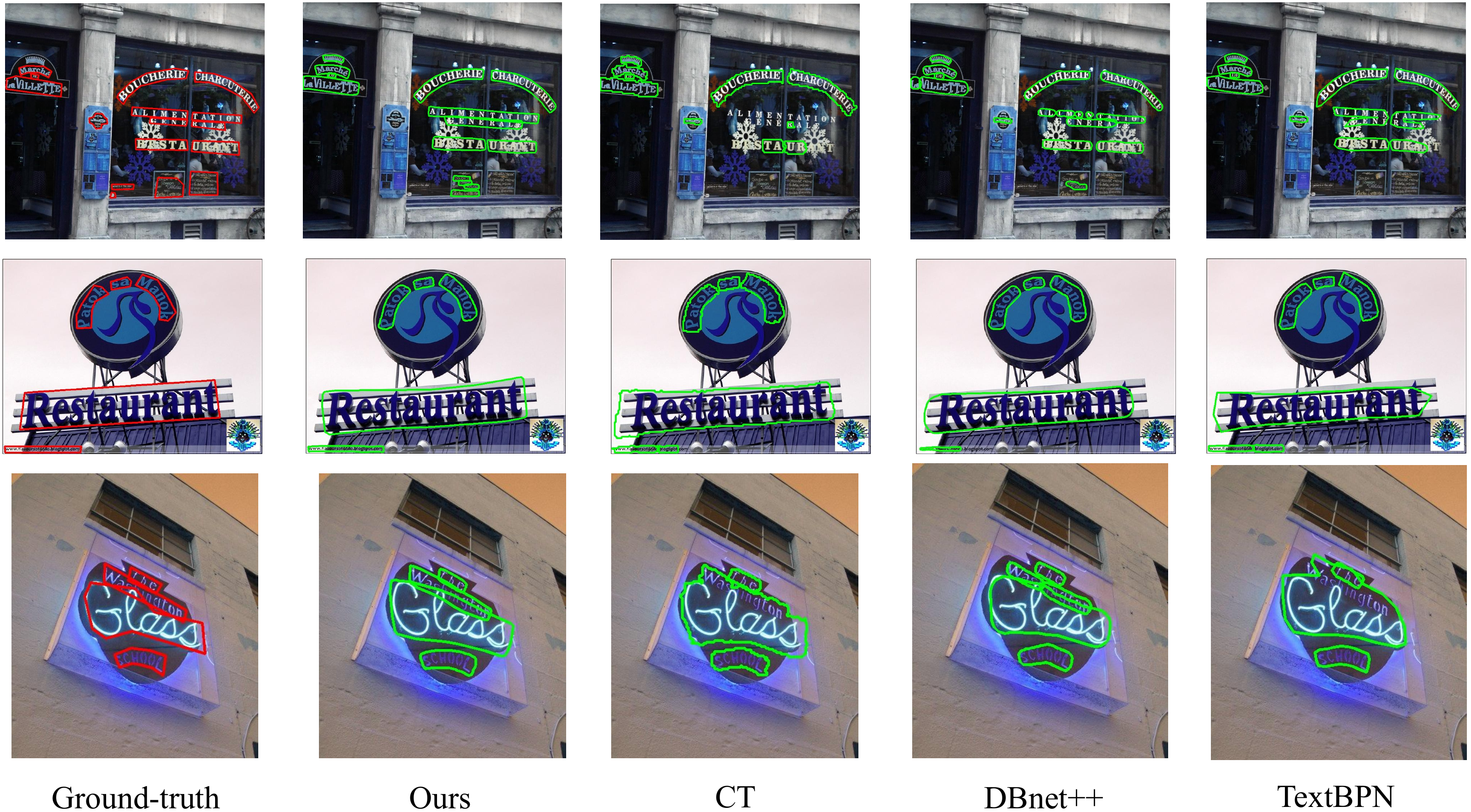} %[图片大小]{图片路径}
   \end{center}
   \vspace{-0.5cm}
    \caption{The visualized detection results obtained by our method, CT~\cite{CentripetalText}, DBNet++~\cite{db++}, and TextBPN\cite{TextBPN} on some selected challenging samples of Total-Text test set.}
   \label{compareresults}
   \vspace{-0.5cm}
   \end{figure*}
 
 \subsection{Implementation details}
 \label{details}
 FPN~\cite{FPN} with ResNet~\cite{resnet} is used as our backbone. Following previous works~\cite{PSE,PAN,DRRG,wang2021pan++}, we first pre-train our models on external text datasets (e.g., SynthText and MLT-2017).
 % and pre-trained on SynthText~\cite{AnkushGupta2016SyntheticDF} for 10 epochs with a fixed learning rate of  $1\times10^{-3}$. Recently, ICDAR-MLT 2017 has widely used as pre-train data in many methods(\eg,\cite{liu2021abcnetv2,DRRG,wang2021pan++}), so we include MLT 2017 as our pre-train data to pre-train our model for anthor 100 epoches based on SynthText pre-trained model using the same strategy.
 % Then, we fine-tune the pre-trained models on several related real-world datasets. For fine-tuning, We follow a poly learning rate policy~\cite{FPN} where the initial learning rate is set to $2\times10^{-4}$ and power is set to 0.9. All models are optimized by the Adam optimizer with the batch size of 16 on two RTX 3090 GPUs.
 Following a poly learning rate policy\cite{FPN}, we fine-tune the pre-trained models on several related real-world datasets with the initial learning rate $2\times10^{-4}$ and the power 0.9. All models are optimized by the Adam optimizer with the batch size of 16 on two RTX 3090 GPUs.
 
 To make fair comparisons in inference speed between different methods, we set up a platform to test some top-performance detectors with the same strategy. Our system consists of one RTX-3090 GPU and one Ryzen-5700X CPU.
 Considering that the performance of the model will affect the time consumption of post-processing, we use official weights provided instead of training them by ourselves. All tests are processed in a single thread with a batch size of 1.
 More details can be referred to Appendix.

 \subsection{Ablation study}
 \textbf{Discussion of Contour Deformation Strategy}:
 % Recent segmentation-based text detection methods like~\cite{PAN,CentripetalText,db,db++} shared a similar pipeline in locating the text kernel. During inference, DB~\cite{db} dropped additional modules to get a faster inference speed. Referring to their settings, we conduct ablation experiments to compare dilating text kernels using Vatti clipping algorithm~\cite{BalaRVatti1992AGS} with our DKE module.
 Recent segmentation-based text detection methods, like~\cite{PAN,CentripetalText,db,db++}, share a similar pipeline in locating the text kernel. DB~\cite{db} leverages the Vatti clipping algorithm~\cite{BalaRVatti1992AGS} as a simple post-processing method to generate the final detection contour based on segmentation results. The main merit of this method is efficiency, but it greatly relies on the accuracy of segmentation results, neglecting adaptively adjusting detected contours. 
 We conduct several experiments under this setting to compare with our proposed contour deformation module DCE, which employs a network for learning the deformation from the kernel boundary to the final detection boundary. According to the results tabulated in Tab.\ref{abs}, DCE obtains 2.9\% and 1.5\% performance gains over baseline in F-measure on Total-Text and CTW1500, respectively, only with a marginal drop in FPS. These results imply that it is worthwhile to incorporate the contour deformation part together with the kernel segmentation for optimization. And learnable kernel expansion module enables outputting more accurate and robust detection boundaries than expanding kernel in a fixed manner. Moreover, our proposed optimal bipartite graph matching loss contributes additional 0.9\% and 1.0\% performance gains in F-measure on Total-Text and CTW1500, respectively, while maintaining the same detection efficiency. This phenomenon validates the importance of contour deformation loss in optimization. The contour deformation losses will be discussed in detail in the next section.
   \begin{table}[t]
    \centering
    % \captionsetup{labelformat=empty}
    % \begin{minipage}{0.6\textwidth}
       \centering
       \small
       \makeatletter\def\@captype{table}\makeatother
       % \caption{}
       % \label{}
       \begin{tabular}{l|ccccl}
          Datasets                    & Method   & P                                 & R                                 & F                                 & FPS          \\
          \hline
          \multirow{3}{*}{Total-Text} & Baseline & 89.3                              & 79.5                              & 84.1                              & \textbf{46}  \\
                                      & +DCE     & 89.9                              & 84.5                              & 87.0                              & 41           \\
                                      & +DCE*     & \multicolumn{1}{l}{\textbf{90.8}} & \multicolumn{1}{l}{\textbf{85.1}} & \multicolumn{1}{l}{\textbf{87.9}} & 41           \\
          \hline
          \multirow{3}{*}{CTW1500}    & Baseline & 84.0                              & 83.0                              & 83.5                              & \textbf{47}  \\
                                      & +DCE     & \textbf{87.4}                     & 82.6                              & 84.9                              & 41           \\
                                      & +DCE*     & \multicolumn{1}{l}{86.6}          & \multicolumn{1}{l}{\textbf{85.2}} & \multicolumn{1}{l}{\textbf{85.9}} & 41
          \end{tabular}
 % 	\end{minipage}\quad
 % \begin{minipage}{0.2\textwidth}
 % 		\centering
 % 		\small
 % 		\makeatletter\def\@captype{table}\makeatother
 % 		\caption{}
 % 		\label{}
 % 		\begin{tabular}{l|cc|cc}
 %          & \multicolumn{2}{c|}{Total-Text} & \multicolumn{2}{c}{CTW1500}  \\
 %    Method   & F    & FPS                      & F    & FPS                   \\
 %    +OTC     & 87.9 & 41                       & 85.8 & 41                    \\
 %    +OTC MI      & 87.3 & 36                       & 85.2 & 37
 %    \end{tabular}
 % 	\end{minipage}
    \vspace{-0.1cm}
    \caption{Ablation study of different contour deformation strategies. Baseline indicates the contour deformation based on Vatti clipping algorithm. DCE indicates our contour deformation named deformable contour expansion, which adopts the simple direct matching loss. DCE* indicates the DCE version, which adopts our proposed optimal bipartite graph matching loss. P is the precision, R is the recall, F is the F-measure, and FPS is the frames per second. }
    \label{abs}
    \vspace{-0.3cm}
 \end{table}
 
  \begin{figure}[t]
    \centering
    % \ref{fig:1}
    % \caption{}
    \includegraphics[width=8.5cm]{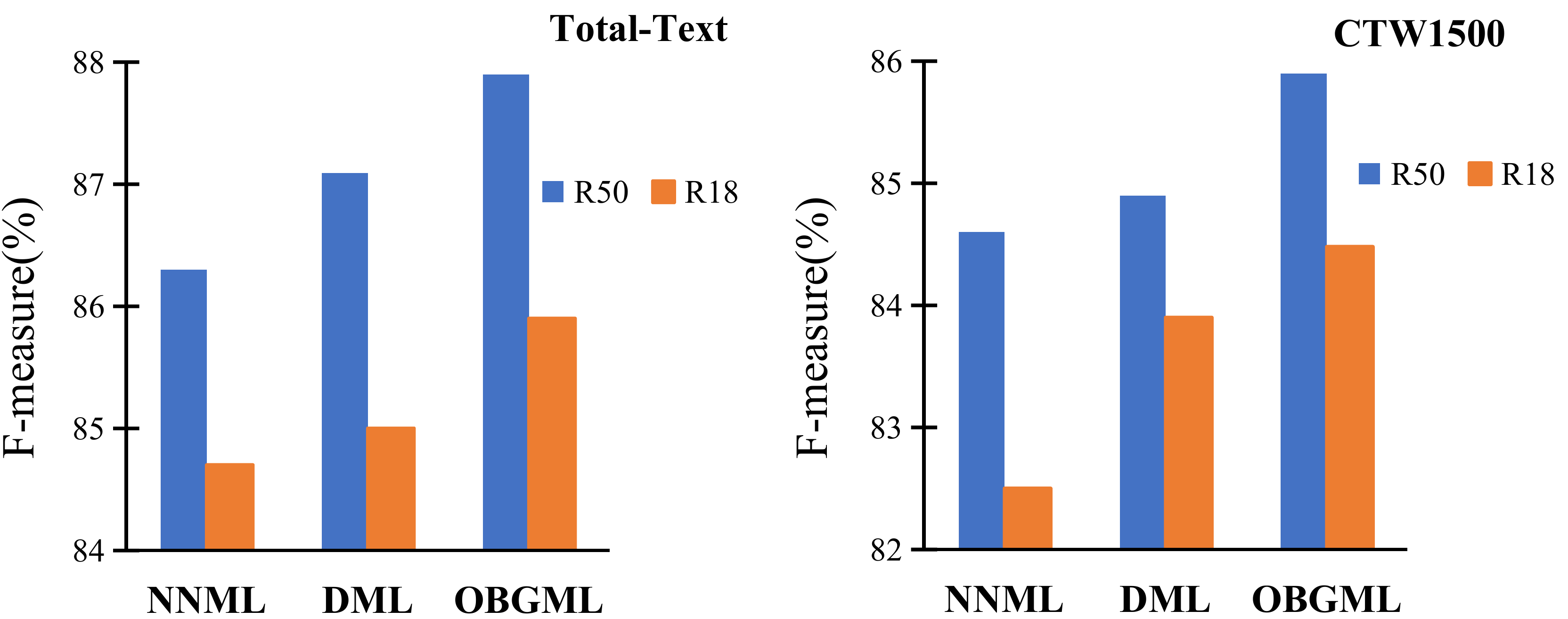} %[图片大小]{图片路径}
    \vspace{-0.5cm}
    \caption{The performances of our method via using different contour deformation losses. R50, R18 denote ResNet50 and ResNet18, respectively.}
    \label{loss_pic}
    \vspace{-0.3cm}
 \end{figure}

 \begin{figure}[b] %插入图片
    \centering %图片居中
    \vspace{-0.5cm}
    \includegraphics[width=8.5cm]{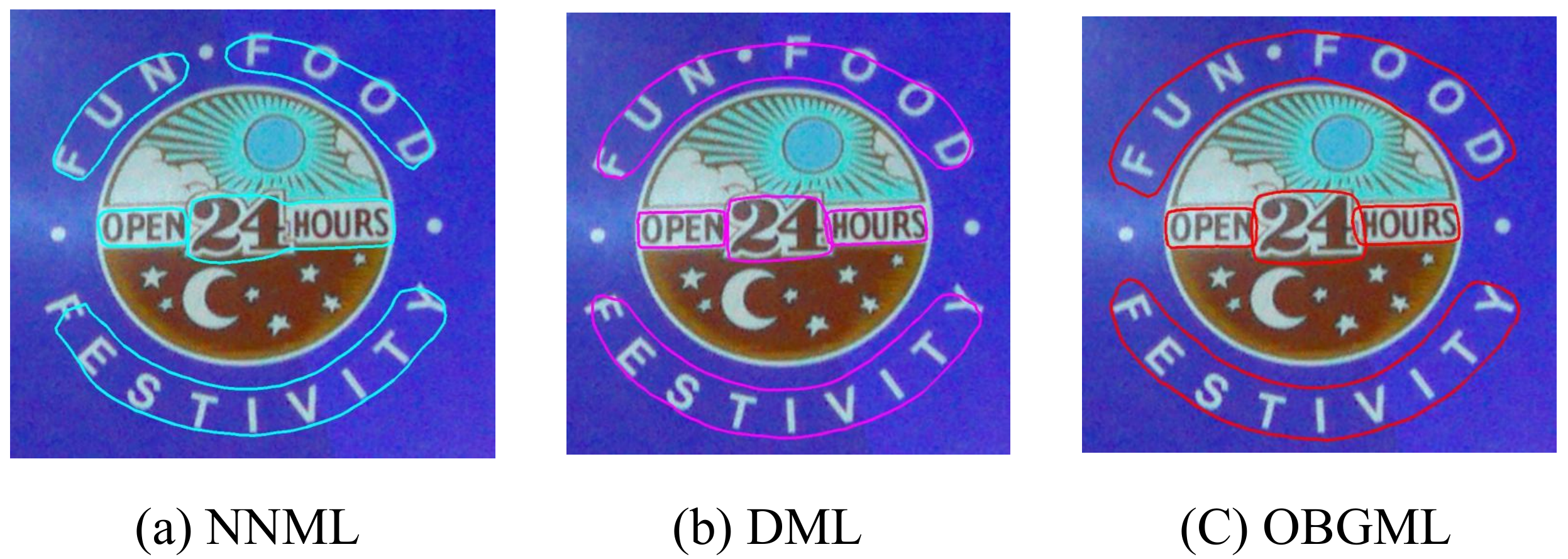} %[图片大小]{图片路径}
    \vspace{-0.7cm}
    \caption{The visualized detection examples with using different contour deformation losses.}
    \label{loss_visual}
 \end{figure}

 \textbf{Discussion of Contour Deformation Loss}:
 We have employed three contour deformation losses, namely Direct Matching Loss (DML), Nearest Neighbor Matching Loss (NNML), and Optimal Bipartite Graph Matching Loss (OBGML), to our Deformable Kernel Expansion (DKE) model. DML does not need to conduct any further vertex pairing. The latter two losses apply different vertex pairing strategies to calculate loss. NNML employs the nearest neighbor search strategy to find the corresponding vertex in ground truth for each predicted vertex. While our proposed OBGML pairs vertex sequences by considering the vertex pairing issue as a bipartite graph matching problem for solution, which guarantees the global optimum. Fig.\ref{loss_pic} demonstrates the performances in F-measure under the same settings. The results show that OBGML outperforms the other two losses on several settings and datasets. We attribute this performance to the superior bipartite graph matching-based vertex pairing strategy. Surprisingly, NNML performs much worse than the most simple and common loss DML when applied for kernel expansion. We attribute its fail to the fact that there is a big difference between our practical scenario and the one designed for NNML. In the process of expansion, predicted vertices are optimized towards the longer side of the text region, which caused the lossing of key ground-truth vertices and complete shape information as shown in Fig.\ref{loss_show}.

 % not all ground truth vertices are fully exploited in the optimization since some predicted vertices may share the same nearest neighbor, while there is no correspondence for some ground truth vertices as shown in Fig.\ref{loss_show}.
 
 %We further study the impact of different deformation strategies to find the best deformation path for transforming text kernels to text boundaries.
 %Fig.\ref{loss_pic} demonstrates the performance comparison of different strategies on different datasets and backbones.
 %When adjusting vertexes with the least cost, the Nearest Neighbor Matching Loss (NML) urges the model to deform the vertexes in kernel contour towards the long-side of the text boundary. As shown in Fig.\ref{loss_show}, this leads to the loss of complete contour information.
 %With R50, NML achieved 86.3\% and 84.6\% in terms of F-measure on Total-Text and CTW1500 datasets, respectively. Replacing NML with DML brings improvements by 0.8\% and 0.3\%. DML maintains the structure of the text region but also greatly increases the cost to transform. Besides, the directing pairing strategy produces some relatively long-distance pairings that are hard to optimize.
 %
 %Those issues are alleviated when we adopt OBGM, which notably reduces cost while retaining contour information. Finally, the F-measure is improved by 1.6\%, 1.3\% on Total-Text and CTW1500 datasets.
 %As shown in Fig.\ref{loss_visual}, the model supervised by OBGM predicts the best detection boundary.
 
 \begin{table}[t]
    \centering
    % \captionsetup{labelformat=empty}
    
    % \label{total}
   
    \begin{tabular}{l|ccccl}
       Datasets                    & Iter    & P    & R    & F    & FPS  \\
       \hline
       \multirow{2}{*}{Total-Text} & Iter. 1 & \textbf{90.8} & \textbf{85.1} & \textbf{87.9} & \textbf{41}   \\
                                   & Iter. 2 & 90.2 & 84.6 & 87.3 & 36   \\
       \hline
       \multirow{2}{*}{CTW1500}    & Iter. 1 & \textbf{86.6} & \textbf{85.2} & \textbf{85.9} & \textbf{41}   \\
                                   & Iter. 2 & 86.4 & 84.1 & 85.2 & 37
       \end{tabular}
      \vspace{-0.2cm}
     \caption{The results of our method in each iteration.}
        \vspace{-0.3cm}
        \label{iter}
    \end{table}

 \begin{table*}[t]
 \centering
 % \caption{labelformat=empty}
 % \label{td500}
 \small
 \begin{tabular}{lccccc|ccc|cccc}
            &                       &               & \multicolumn{3}{c|}{TD500}                                                                                        & \multicolumn{3}{c|}{CTW1500}                                                                                                         & \multicolumn{4}{c}{Total-Text}                                                                                                                                                    \\
 Method     & Venue                 & Backbone      & P                        & R                                         & F                                          & P                                         & R                                          & F                                           & P                                          & R                                         & F                                          & FPS                                         \\
 \hline
 PAN~\cite{PAN}        & ICCV'2019             & R18           & 84.4                     & \multicolumn{1}{l}{\textcolor{red}{83.8}} & \multicolumn{1}{l|}{84.1}                  & \multicolumn{1}{l}{86.4}                  & \multicolumn{1}{l}{81.2}                   & \multicolumn{1}{l|}{83.7}                   & \multicolumn{1}{l}{\textcolor{blue}{89.3}} & \multicolumn{1}{l}{81.0}                  & \multicolumn{1}{l}{85.0}                   & -                                           \\
 CT~\cite{CentripetalText}         & NeurIPS'2021          & R18           & \textcolor{blue}{90.0}   & \multicolumn{1}{l}{82.5}                  & \multicolumn{1}{l|}{\textcolor{red}{86.1}} & \multicolumn{1}{l}{\textcolor{red}{88.3}} & \multicolumn{1}{l}{79.9}                   & \multicolumn{1}{l|}{\textcolor{blue}{83.9}} & \multicolumn{1}{l}{\textcolor{red}{90.5}}  & \multicolumn{1}{l}{\textcolor{blue}{82.5}}                  & \multicolumn{1}{l}{\textcolor{red}{86.3}}  & \multicolumn{1}{l}{47.3}                    \\
 DBNet~\cite{db}      & AAAI'2020             & R18$^\dagger$ & \textcolor{red}{90.4}    & \multicolumn{1}{l}{76.3}                  & \multicolumn{1}{l|}{82.8}                  & \multicolumn{1}{l}{84.8}                  & \multicolumn{1}{l}{77.5}                   & \multicolumn{1}{l|}{81.0}                   & \multicolumn{1}{l}{88.3}                   & \multicolumn{1}{l}{77.9}                  & \multicolumn{1}{l}{82.8}                   & \multicolumn{1}{l}{\textcolor{red}{81.1}}   \\
 DBNet++~\cite{db++}    & TPAMI'2022            & R18$^\dagger$ & \multicolumn{1}{l}{87.9} & \multicolumn{1}{l}{82.5}                  & \multicolumn{1}{l|}{85.1}                  & \multicolumn{1}{l}{86.7}                  & \multicolumn{1}{l}{\textcolor{blue}{81.3}} & \multicolumn{1}{l|}{\textcolor{blue}{83.9}} & \multicolumn{1}{l}{87.4}                   & \multicolumn{1}{l}{79.6}                  & \multicolumn{1}{l}{83.3}                   & \multicolumn{1}{l}{\textcolor{blue}{75.1}}  \\
 \hline
 \textbf{DKE(Ours)}      & \multicolumn{1}{c}{-} & R18           & 87.9                     & \textcolor{blue}{83.1}                    & \textcolor{blue}{85.4}                     & \textcolor{blue}{86.9}                    & \textcolor{red}{82.2}                      & \textcolor{red}{84.5}                       & 88.2                                       & \textcolor{red}{83.7}                     & \textcolor{blue}{85.9}                     & 67.2                                        \\
 % \hline
 \hline
 SAE~\cite{SAE}        & CVPR'2019             & R50           & 84.2                     & 81.7                                      & 82.9                                       & 82.7                                      & 77.8                                       & 80.1                                        & -                                          & -                                         & -                                          & -                                           \\
 PSENet-1s~\cite{PSE}  & CVPR'2019             & R50           & -                     & -                                     & -                                       & 82.5                                    & 79.9                                       & 81.2                                       & -                                       & -                                   & -                                       & -                                           \\
 SPCNet~\cite{SPCnet}     & AAAI'2019             & R50           & -                        & -                                         & -                     & -                                         & -                                          & -                       & 83.0                                       & 82.8                                      & 82.9                                       & -                                           \\
 CounterNet~\cite{contournet} & CVPR'2020             & R50           & -                        & -                                         & -                                          & 83.7                                      & \textcolor{blue}{84.1}                     & 83.9                                        & 86.9                                       & 83.9                                      & 85.4                                       & -                                           \\
 DBNet~\cite{db}      & AAAI'2020             & R50$^\dagger$ & 91.5                     & 79.2                                      & 84.9                                      & 86.9                                      & 80.2                                       & 83.4                                        & 87.1                                       & 82.5                                      & 84.7                                       & \textcolor{red}{42.2}                       \\
 FCENet~\cite{FCE}     & CVPR'2021             & R50$^\dagger$ & -                        & -                                         & -                                          & 87.6                                      & 83.4                                       & \textcolor{blue}{85.5}                      & 89.3                                       & 82.5                                      & 85.8                                       & -                                           \\
 TextBPN~\cite{TextBPN}    & ICCV'2021             & R50           & 86.6                     & 84.5                                      & 85.6                                       & 86.5                                      & 83.6                                       & 85.0                                        & \textcolor{blue}{90.7}                     & \textcolor{blue}{85.2}                    & \textcolor{blue}{87.9}                     & 17.2                                        \\
 DBNet++~\cite{db++}    & TPAMI'2022            & R50$^\dagger$ & 91.5                     & 83.3                                      & 87.2                                       & \textcolor{blue}{87.9}                    & 82.8                                       & 85.3                                        & 88.9                                       & 83.2                                      & 86.0                                       & 40.8                                        \\
 FewNet~\cite{Fewnet}     & CVPR'2022             & R50           & \textcolor{blue}{91.6}   & \multicolumn{1}{l}{\textcolor{blue}{84.8}}                 & \multicolumn{1}{l|}{\textcolor{blue}{88.1}}                  & \multicolumn{1}{l}{\textcolor{red}{88.1}} & \multicolumn{1}{l}{82.4}                   & \multicolumn{1}{l|}{85.2}                   & \multicolumn{1}{l}{\textcolor{blue}{90.7}}                   & \multicolumn{1}{l}{\textcolor{red}{85.7}} & \multicolumn{1}{l}{\textcolor{red}{88.1}}  & -                                           \\
 \hline
 \textbf{DKE(Ours)}     & \multicolumn{1}{c}{-} & R50           & \textcolor{red}{91.7}    & \textcolor{red}{85.9}                     & \textcolor{red}{88.7}                      & 86.6                                      & \multicolumn{1}{l}{\textcolor{red}{85.2}}  & \multicolumn{1}{l|}{\textcolor{red}{85.9}}  & \multicolumn{1}{l}{\textcolor{red}{90.8}}  & \multicolumn{1}{l}{85.1}                  & \multicolumn{1}{l}{\textcolor{blue}{87.9}} & \textcolor{blue}{41.0}
 \end{tabular}
 \vspace{-0.2cm}
  \caption{Detection results on the MSRA-TD500, CTW1500, and Total-Text datasets. ${\dagger}$ indicates that the backbones have adopted the modulated deformable convolutions~\cite{zhu2019deformable}. The best and second-best F-measures are highlighted in \textcolor{red}{red} and \textcolor{blue}{blue}, respectively.
    }
    \label{threedatasets}
    \vspace{-0.3cm}
 \end{table*}
 
 \textbf{Iterations for Contour Deformation}: Traditional contour-based scene text detection methods\cite{weaklyarbitrary,TextBPN,dai2021progressive} often need to iteratively adjust the coarse predicted boundaries several times. The initial inputs predicted by them often target at text region boundaries. Different from them, DCE module applies the kernel expansion to model the contour deformation.
The local features on text boundaries are less reliable than the ones on the kernel boundary, which are located at the more central area of the text region. The reliable features speed up the optimization process, and only a few iterations of contour adjustment can guarantee good convergence. The observations in Tab.\ref{iter} well validate this. Tab.\ref{iter} reports the performances of our model in the first two iterations on Total-text and CTW1500 datasets. The results reveal that our model can obtain an outstanding performance in the first iteration, but more iterations degenerate the performances.

 \begin{table}[b]
   \vspace{-0.2cm}
  \small
 \centering

 \begin{tabular}{lcccc}
 Method     & Venue                         & P                        & R                        & F                         \\
 \hline
 Seglink~\cite{seglink}    & CVPR'2017                     & 73.1                     & 76.8                     & 75.0                      \\
 TextSnake~\cite{textsnake}  & ECCV'2018                     & 84.9                     & 80.4                     & 82.6                      \\
 SAE~\cite{SAE}        & CVPR'2019                     & 88.3                     & 85.0                     & 86.6                      \\
 PSENet-1s~\cite{SAE}        & CVPR'2019                     & 88.7                    & 85.5                     & 87.1                      \\
 PAN~\cite{PAN}        & ICCV'2019                     & 84.0                     & 81.9                     & 82.9                      \\
 SPCNet~\cite{SPCnet}  & AAAI'2019                     & 88.7                     & 85.8                     & 87.2                      \\
 CounterNet~\cite{contournet} & CVPR'2020                     & 87.6                     & \textcolor{blue}{86.1}    & 86.9                      \\
 DBNet~\cite{db}      & AAAI'2020                     & \textcolor{red}{91.8}    & 83.2                     & 87.3                      \\
 DRRG~\cite{DRRG}       & CVPR'2020                     & 88.5                     & 84.6   & 86.5                      \\
 FCENet~\cite{FCE}     & \multicolumn{1}{l}{CVPR'2021} & \multicolumn{1}{l}{90.1} & \multicolumn{1}{l}{82.6} & \multicolumn{1}{l}{86.2}  \\
 DBNet++~\cite{db++}    & TPAMI'2022                    & 90.9                     & 83.9                     & 87.3                      \\
 FewNet~\cite{Fewnet}     & CVPR'2022                     & 90.9                     & \textcolor{red}{87.3}                     & \textcolor{red}{89.1}     \\
 \hline
\textbf{DKE(Ours)}      & -                             & \textcolor{blue}{91.0}   & 84.3                     & \textcolor{blue}{87.5}
 \end{tabular}
%  \vspace{-0.2cm}
 \caption{Detection results on the ICDAR2015 dataset. }
       \label{icdar2015}
 \end{table}

 \subsection{Comparisons with Previous Methods}
 \textbf{Curved text detection.}
 Tab.\ref{threedatasets} reports the scene text detection performances on two curved text datasets named Total-Text and CTW1500. Our approach achieves the best and the second-best performances in F-measure on CTW500 and Totat-Text, respectively. For example, compared with DBNet++, which is known as one of the most influential and efficient segmentation-based approaches, our method almost enjoys the same efficiency. However, our method outperforms it by 2.6\% and 1.9\% in F-measure on Total-Text when the backbone is ResNet18 and ResNet50, respectively. This gain on CTW500 is 0.6\%. CT is also a segmentation-based approach and performs very well on all two datasets when the backbone is ResNet18. However, it is worthwhile to point out that our method enables us to achieve similar performance with only 70\% of its inference time. TextBPN is the best-performed contour-based approach among all baselines. Our method achieves the similar or even better performance in F-measure but enjoys \textcolor{blue}{2.4} times faster inference speed. Clearly, our method enjoys a good tradeoff between detection performance and inference speed. Fig.\ref{compareresults} visualizes some detection results of several top-performance detectors on the Total-Text dataset. These visualizations also confirm that our method can provide more accurate detection results and better separate two adjacent text regions. 
 
 \textbf{Quadrangular Text detection.}
 According to the results in Tab.\ref{threedatasets} and Tab.\ref{icdar2015}, similar phenomena can be observed on two quadrangular text datasets named MSRA-TD500 and ICDAR2015. Our method still achieves competitive performances on these two benchmarks. For example, our method performs best in F-measure on MSRA-TD500 when the backbone is ResNet50. FewNet\cite{Fewnet} is the recent SOTA detector. The performance gain of our method over it is 0.8\% on MSRA-TD500. 
 
 \section{Conclusion}
 In this paper, we present an effective arbitrary-shaped scene text detector named Deformable Kernel Expansion (DKE), which inherits merits of both segmentation and contour based methods. 
 %DKE adopts a lightweight segmentation network to segment a shrunken text area as the text kernel for offer an initial detection boundary, and employs a regression network to expand the text kernel contour to the final detection boundary with only one iteration of optimization. DKE can inherits good text shape modelling 
 DKE takes advantage of the arbitrary-shaped text region modeling power of segmentation-based detectors at the pixel level by segmenting text kernels as the initial boundaries. Those kernels can be deemed as shrunken versions of the text regions. Based on the text kernel, DKE learns to expand the text kernel rather than adjust the text boundary several times as the conventional contour-based scene text detectors do. Such an expansion enables it to be optimized in only one iteration, which highly speeds up the inference and also avoids the complicated pixel-level post-processing adopted by most of the segmentation-based detectors. Extensive experimental results on several benchmarks validate the effectiveness of DKE and verify that our method can achieve a good tradeoff between accuracy and efficiency.
 
 % keeping the real-time inference speed
 %\textbf{Limitation} One limitation of our method is that it can not deal with
 %cases like "text inside text" and "interlaced text" due to the segmentation-based detection head. This limitation is common for segmentation-based detectors.

 %%%%%%%%% REFERENCES
 {\small
 \bibliographystyle{ieee_fullname}
 \bibliography{egbib}
 }
 
 \end{document}